\documentclass[11pt]{article} 
\usepackage{url}
\usepackage{graphicx} 
\usepackage{epstopdf}
\usepackage[colorlinks, linkcolor=blue, anchorcolor=blue, citecolor=blue]{hyperref}
\usepackage[margin=1in]{geometry}
\usepackage[normalem]{ulem}
\usepackage[export]{adjustbox} 
\usepackage{mathtools, cuted}
\usepackage{enumerate}
\usepackage{enumitem}
\usepackage{microtype}
\usepackage{graphicx}
\usepackage{booktabs}
\usepackage{multirow} 
\usepackage{makecell}
\usepackage{amsmath}
\usepackage{amssymb}
\usepackage{mathtools}
\usepackage{amsthm}
\usepackage{algorithm}
\usepackage{algorithmic}
\usepackage{natbib}
\usepackage{listings}
\usepackage{subfigure}

\usepackage{kpfonts}
\DeclareMathAlphabet{\mathsf}{OT1}{cmss}{m}{n}

\SetMathAlphabet{\mathsf}{bold}{OT1}{cmss}{bx}{n}

\usepackage{amsmath}
\usepackage{amssymb}
\usepackage{mathtools}
\usepackage{amsthm}

\usepackage[capitalize,noabbrev]{cleveref}

\theoremstyle{plain}

\theoremstyle{definition}

\theoremstyle{remark}

\usepackage[textsize=tiny]{todonotes}

\author{Zichong Li, Xinyu Feng, Yuheng Cai, Zixuan Zhang, Tianyi Liu, \\Chen Liang, Weizhu Chen, Haoyu Wang, Tuo Zhao \footnote{Li, Feng, Cai, Zhang and Zhao are affiliated with Georgia Tech. Liang, Chen are affiliated with Microsoft Azure. Liu is affiliated with Amazon. Wang  is affiliated with University at Albany. Correspondence to \url{zli911@gatech.edu} and \url{tourzhao@gatech.edu}.}}

\title{LLMs Can Generate a Better Answer by Aggregating Their Own Responses}
\date{}

\begin{document}

\maketitle

\begin{abstract}

Large Language Models (LLMs) have shown remarkable capabilities across tasks, yet they often require additional prompting techniques when facing complex problems. While approaches like self-correction and response selection have emerged as popular solutions, recent studies have shown these methods perform poorly when relying on the LLM itself to provide feedback or selection criteria. We argue this limitation stems from the fact that common LLM post-training procedures lack explicit supervision for discriminative judgment tasks.
In this paper, we propose Generative Self-Aggregation (GSA), a novel prompting method that improves answer quality without requiring the model's discriminative capabilities. GSA first samples multiple diverse responses from the LLM, then aggregates them to obtain an improved solution. Unlike previous approaches, our method does not require the LLM to correct errors or compare response quality; instead, it leverages the model's generative abilities to synthesize a new response based on the context of multiple samples.
While GSA shares similarities with the self-consistency (SC) approach for response aggregation, SC requires specific verifiable tokens to enable majority voting. In contrast, our approach is more general and can be applied to open-ended tasks. Empirical evaluation demonstrates that GSA effectively improves response quality across various tasks, including mathematical reasoning, knowledge-based problems, and open-ended generation tasks such as code synthesis and conversational responses.
\end{abstract}

\section{Introduction}

Large Language Models (LLMs) have demonstrated remarkable capabilities across a wide range of tasks, yet they often struggle with complex problems requiring careful deliberation or multi-step reasoning \cite{chatgpt}. This limitation has prompted the development of various prompting techniques, with self-correction \cite{self-correct-survey, selfrefine} and choose-from-N \cite{constitutionalAI, scale_test, choose-from-N} emerging as popular approaches. In self-correction, models revise their initial responses based on feedback, while choose-from-N methods generate multiple candidates and select the best one. However, recent studies have shown that these methods perform suboptimally when relying on the LLM itself to provide feedback or selection criteria without external guidance \cite{cannotcorrect, generative_reward}.

\begin{figure*}[tb!]
    \centering
    \includegraphics[width=\linewidth]{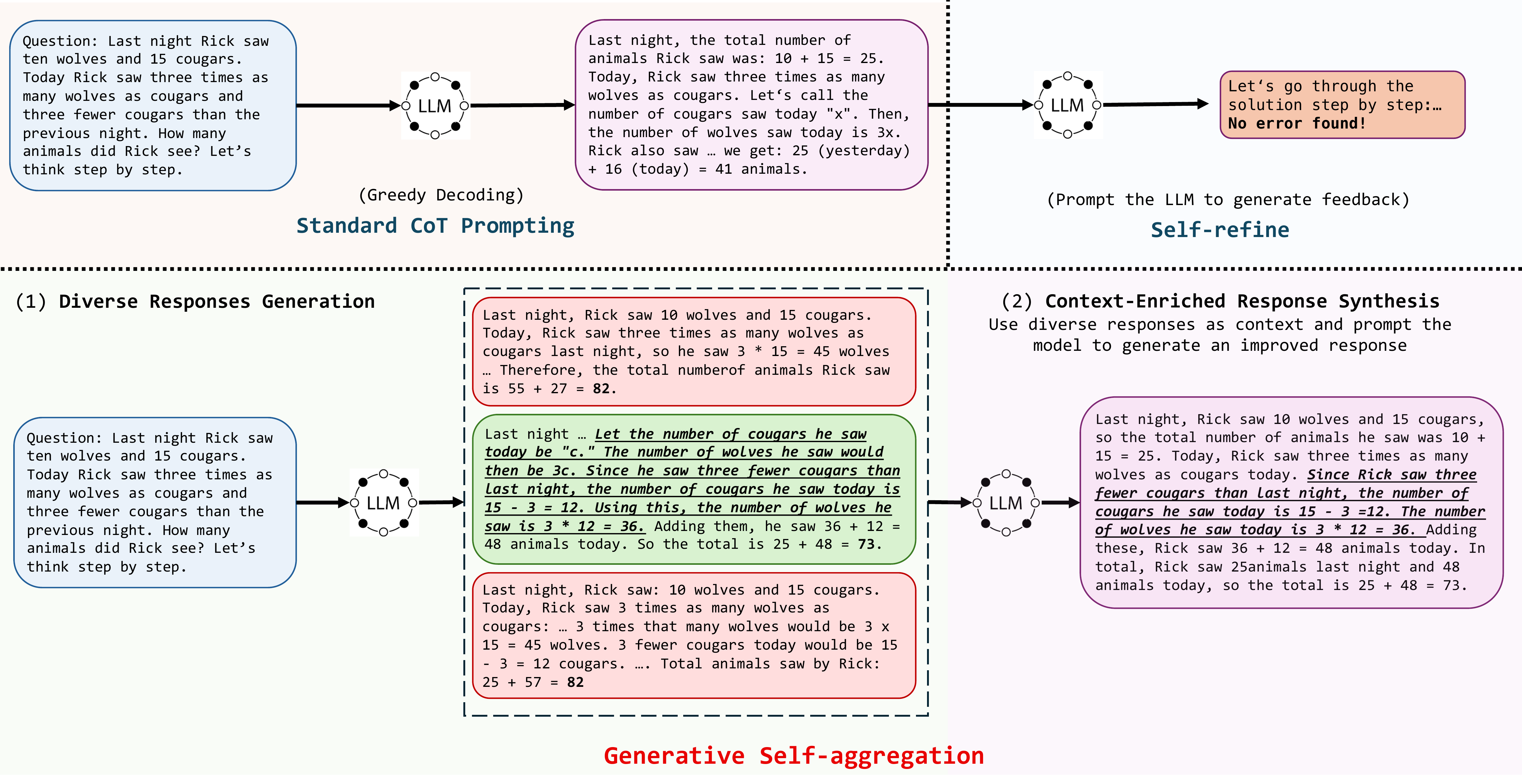}
    \vspace{-0.1in}
    \caption{Illustration of Generative Self-aggregation with an example on a math problem using Llama 3 8B as the language model.}
    \label{fig:main}
\end{figure*}

We argue that these limitations stem from a fundamental characteristic of LLM training: while standard training enables impressive zero-shot text generation capabilities, it lacks explicit supervision for discriminative judgment tasks. This hypothesis is supported by recent work on generative reward models \cite{generative_reward}, which demonstrates that LLMs' judgment capabilities can be significantly enhanced through specialized training. However, such training approaches necessitate extensive data curation efforts and considerable computational resources for fine-tuning, making them impractical for many applications.

To the best of our knowledge, existing methods that successfully self-improve without additional training on discriminative processes fall into two main categories: approaches utilizing external feedback (e.g., code executors \cite{external_code1} or computational verifiers \cite{external_symbolic}) and the self-consistency method \cite{selfconsistency}. The external feedback approach requires careful task-specific design and is not universally applicable across different domains. 
Self-consistency takes a different approach by generating multiple samples and employing majority voting among them. However, its applicability is limited to tasks with specific, verifiable tokens as answers, such as mathematical reasoning or multiple-choice questions, making it unsuitable for open-ended tasks. Moreover, self-consistency can only aggregate final answers, overlooking the valuable reasoning processes behind these answers.

Building on these insights, we propose a novel method, Generative Self-Aggregation (GSA), that improves answer quality by aggregating information from multiple responses. Unlike previous methods, GSA does not require the model to explicitly judge or compare responses and does not need additional training.
As illustrated in Figure \ref{fig:main}, GSA first generates multiple diverse responses, then uses these responses as context to prompt the model to aggregate them and generate an improved solution.
The common LLMs' training paradigm of predicting subsequent tokens based on input context allows the model to identify and learn from stronger solutions through its natural text generation capabilities, enabling the model to combine strengths from different solutions. Unlike traditional self-consistency which relies on majority voting, our approach leverages the generative power of LLM for aggregating multiple responses. By utilizing the reasoning process rather than just the final answers, the model has access to more information that can be aggregated to improve the final solution. Moreover, this generative aggregation approach extends beyond specific-answer tasks to open-ended problems where majority voting would not be applicable.

Through extensive experimentation across mathematical reasoning, knowledge-based, and open-ended generation tasks, we demonstrate that our method outperforms both self-correction and choose-from-N baselines across different tasks and model scales. Our method achieves comparable or better performance to self-consistency on tasks with verifiable answers, while self-consistency cannot be applied to open-ended tasks. Our ablation studies demonstrate the method's robustness across different sampling strategies. Further analysis of likelihood distributions reveals that LLMs are more confidence on generating new responses than selecting among existing ones, providing empirical support for our framework.

The rest of the paper is organized as follows: Section 2 reviews related work in LLM prompting techniques; Section 3 introduces our proposed Generative Self-Aggregation approach; Section 4 presents comprehensive experimental results;  Section 5 draws a brief conclusion and discusses potential future directions.

\section{Related work}

\subsection{Self-consistency}
Self-consistency \cite{selfconsistency} is a decoding strategy that improves Chain-of-Thought (CoT) prompting by leveraging multiple reasoning paths. Instead of using greedy decoding to generate a single solution, it first samples multiple diverse reasoning paths and then aggregates their final answers through majority voting to determine the most consistent one. Its success on mathematical reasoning and common sense questions demonstrates LLMs' ability to generate correct solutions across different attempts. However, its application to open-ended tasks remains challenging due to the lack of clear voting mechanisms.

\subsection{Self-correction}

Self-correction in Large Language Models has emerged as a significant research direction, focusing on models' ability to recognize and improve their outputs based on feedback \cite{self-correct-survey, finetune1, selfrefine}.
A substantial body of work explores self-correction with external feedback sources, such as human annotations \cite{external_gt}, code executors \cite{external_code1, external_code2, external_code3}, or symbolic reasoning tools \cite{external_symbolic}. While effective, these approaches are limited by their reliance on additional knowledge sources or tools that may not always be available.
When only the language model itself is available, researchers have proposed various intrinsic self-correction methods. Self-Refine \cite{selfrefine} and RIC prompting \cite{intrinsic_comp} prompt the model to provide feedback on and refine its previous outputs. However, recent studies have shown that these approaches, which rely on LLMs' ability to make discriminative judgments about their own outputs, often produce inaccurate assessments and yield suboptimal results \cite{cannotcorrect}. 

To address these limitations, some researchers have explored training-based approaches to enhance self-correction capabilities. These include supervised fine-tuning methods \cite{finetune1, finetune2} and reinforcement learning approaches \cite{finetune_rl}. While promising, these methods require substantial human-annotated training data, limiting their practical applicability.

\subsection{Choose-from-N Methods}

Another prominent approach to improve language model outputs is the choose-from-N paradigm, where multiple candidate responses are first generated and then selected based on specific selection criteria. Best-of-N sampling \cite{learn_to_sum, reg_bon} represents a widely adopted variant of this approach, utilizing reward model scores as selection criteria during decoding to better align responses with human preferences. This methodology has been extended by works incorporating specialized verifiers or process reward models (PRM, \citet{scale_test}) to enhance selection accuracy. However, the effectiveness of these approaches heavily depends on having well-trained reward models that accurately reflect human preferences.

An alternative strategy employs the language model itself as the evaluators for selection.  
For instance, Constitutional AI \cite{constitutionalAI} introduces Reinforcement Learning from AI Feedback (RLAIF), employing LLMs to identify harmful content and generate preference labels for training.
\citet{mtbench} further establish that strong LLMs can provide judgments that correlates with human preferences.
While this correlation can be effective for tasks like model evaluation or dataset construction where errors can be averaged out, using LLMs as judges for selecting better responses in individual cases may not be optimal.
Moreover, recent studies \cite{arellmasjudge, judging-llmasjudge} have identified several challenges in these approaches.
\citet{generative_reward} shows that LLMs' zero-shot judgments may not always fully align with human preferences.
\citet{judging-llmasjudge} demonstrates that recent open-source LLMs' alignment performance falls considerably short of human-to-human agreement, with their evaluations often deviating significantly from human assessments.
Our method takes an alternative approach without requiring LLMs' discriminative judging capabilities.

\subsection{Multi-model Collaboration}
Recent research has explored the potential of leveraging multiple large language models as interactive agents to solve complex problems. Various multi-agent frameworks have emerged where models assume distinct roles, such as debaters and judges \cite{debate1,debate2}. For example, \citet{debate3} present an iterative discussion framework that enables multiple LLM agents to engage in round-table discussions with confidence-weighted voting. \citet{moa} propose a multi-layer design to iteratively aggregate responses from different models.
However, their fundamental focus differs from ours - they aim to leverage the complementary strengths across heterogeneous models, while our method explores how a single model can improve over its own responses. Moreover, all these multi-agent approaches require simultaneous deployment of multiple models, resulting in substantially higher computational and resource costs.

\section{Methodology}
We propose Generative Self-Aggregation (GSA), a prompting method that improves answer quality without relying on LLMs' discriminative ability.
Our method consists of two key steps: (1) diverse response generation and (2) context-enriched response synthesis. 
Unlike traditional self-consistency, which relies on majority voting to find agreement among multiple final outputs, our approach enables the model to synthesize an improved solution by learning from diverse attempts. Our method operates within the model's generative framework without requiring any discriminative judgments (such as selecting or judging responses) and does not require additional training.

\subsection{Diverse Response Generation}
Given a language model $\mathcal{M}(\cdot)$ and query $q$, we first generate $n$ diverse responses by sampling from the model's distribution:
\begin{align*}
r_i \sim \mathcal{M}(r | q), \quad i = 1,\ldots,n.
\end{align*}
We can employ various sampling strategies, such as temperature sampling \cite{temp_sampling} or nucleus sampling \cite{nucleus_sampling}, to generate these candidates. The diversity of these responses is crucial for providing rich context in the subsequent synthesis step. 
For example, in mathematical reasoning tasks, diverse candidates may explore different solution paths, potentially containing valuable correct intermediate steps even when reaching incorrect final answers.
In knowledge-based tasks, they can access different aspects of the model's internal knowledge that a single deterministic generation might miss.

\subsection{Context-Enriched Response Synthesis}
After generating diverse candidates, we construct an enriched prompt by combining the original task query with the generated responses and ask the model to generate an new response:
\begin{align*}
r' \sim \mathcal{M}(r | \text{Prompt}(q, \{r_i\}_{i=1,...,n}).
\end{align*}
Standard training of next-token prediction enables the language model to attend to and learn from useful content in the provided context.
By providing the model multiple attempts, we enable it to identify effective reasoning patterns or knowledge and combine them into a more refined response, potentially combining the strengths while avoiding their individual weaknesses.
The following box presents a prompt template that we use for a coding task MBPP \cite{mbpp}:
\begin{lstlisting}[language=, caption={}]
### Here is the problem:
{query}
### Reference Solutions:
{diverse_responses}
### Instructions:
1. Review the above solutions.
2. Generate a Python function that solves the Problem.
3. Provide a brief explanation of your reasoning.
4. Ensure your code is enclosed within a ```python``` code block.

\end{lstlisting}

Figure \ref{fig:main} presents an example using Llama 3 8B \cite{llama3} on a mathematical reasoning task. When given this problem, standard zero-shot chain-of-thought prompting with greedy decoding produces an incorrect solution.
Self-refine ask the model to provide a feedback to the generated solution, but failing to correctly identify the errors.
Our method, in contrast, first generates three diverse solutions through sampling, with two arriving at an incorrect answer and one reaching the correct answer.
Instead of attempting to select the final answer through voting, our method provides these diverse attempts as context and prompts the model to generate an improved solution.

The resulting response not only arrives at the correct answer but also demonstrates an evolution in reasoning strategy: the original correct solution first derives the number of wolves as $3c$ with $c$ defined as the number of cougars, and then calculate $c$, following the sequential order of conditions in the question. Differently, the improved solution takes a more direct approach without introducing variables. This transformation suggests that the model can not only identify correct reasoning patterns from the provided attempts but also simplify them into more straightforward solution paths.

\section{Experiments}

Our experimental section starts with the basic setup, followed by our main results and ablation studies. We include discussions of the experimental results along with case studies examining math reasoning and coding problems. Our code is available at \url{https://github.com/zichongli5/Generative-Self-Aggregation}.

\subsection{Experimental Setup}

\textbf{Language models.} We evaluate our method using four different LLMs to ensure robustness across model scales and architectures. We use the instruction-tuned version of the open-sourced models for experiments.

\noindent$\bullet$ \textbf{Llama 3 8B} \cite{llama3} is an open-source model pretrained on around 15 trillion tokens, with improved performance on a wide range of industry benchmarks. 

\noindent$\bullet$ \textbf{Gemma 2 9B} \cite{gemma2} is a mid-sized model in the Gemma 2 family that leverages knowledge distillation during training. It was trained on 8 trillion tokens and incorporates several architectural improvements.

\noindent$\bullet$ \textbf{Qwen 2.5 14B} \cite{qwen2} belongs to the latest series of Qwen LLMs. It achieves significant improvement on knowledge and reasoning upon previous series. 

\noindent$\bullet$ \textbf{GPT4o Mini} is a scaled-down variant of the GPT-4 family, designed to offer faster inference and lower cost without compromising too much on generation quality.
\\

\noindent\textbf{Tasks and datasets.} We conduct comprehensive evaluations across diverse tasks with task-specific evaluation metrics:

\noindent $\bullet$ \textbf{Mathematics reasoning.} For mathematical reasoning capabilities, we use Grade School Math 8K (GSM8K; \citealt{gsm8k}), a dataset of grade school math word problems that require multi-step reasoning. We also include Math \cite{MATH}, a diverse collection of mathematical problems spanning algebra, arithmetic, and geometry. SVAMP \cite{svamp} serves as a challenge set designed for more robust evaluation of models against elementary level math word problems.

\noindent $\bullet$ \textbf{Knowledge Tasks.} For testing knowledge application, we employ the Massive Multitask Language Understanding benchmark (MMLU; \citealt{mmlu}), which covers 57 subjects ranging from mathematics to law.
Due to resource constraints, we randomly sample 10\% of the MMLU test set for evaluation.
Additionally, we use Graduate-Level Google-Proof Q\&A Benchmark (GPQA; \citealt{gpqa}), a benchmark with very hard question written by experts in biology, physics, and chemistry, which also requires strong reasoning ability.

\noindent $\bullet$ \textbf{Open-ended Tasks.} To evaluate performance on less constrained tasks, we use MT-bench \cite{mtbench}, a multi-turn dialogue benchmark specifically designed to assess open-ended conversation capabilities. Alpaca Eval \cite{alpaca_eval} provides a comprehensive suite for evaluating instruction-following abilities. For programming tasks, we include MBPP \cite{mbpp}, which contains 974 basic Python programming problems. 
For MT-bench, we employ GPT-4 \cite{gpt4} as the evaluator to score the response.
For Alpaca eval, we employ GPT-4 to calculate win rate against responses from GPT4-turbo.
\\

\noindent \textbf{Baselines.} We evaluate our approach against the following baselines:

\noindent$\bullet$ \textbf{Greedy}: Standard greedy decoding (temperature = 0).

\noindent$\bullet$ \textbf{Self-Refine} \cite{selfrefine}: Iterative response refinement based on model-generated feedbacks.

\noindent$\bullet$ \textbf{Self-Consistency} \cite{selfconsistency}: Generation of multiple responses followed by majority voting of final answers.

\noindent$\bullet$ \textbf{Choose-from-N}: Generation of multiple responses followed by model selection via index output

\noindent$\bullet$ \textbf{Best-of-N (Oracle)}: Upper bound accuracy where a response is considered correct if any candidate solution is correct. For MT-bench, we report the maximum score among all candidates.
\\

To ensure fair comparison, we standardize computational budget across all methods by fixing the number of model calls to 4 in the main results. For Self-Consistency, we generate four diverse candidates using temperature sampling and apply majority voting on the final answer. We randomly select three of these candidates for our method and choose-from-N. For Self-Refine, we limit the feedback-refinement loop to 2 iterations.
We maintain consistent prompt templates across our method and choose-from-N, while adhering to the original prompts for Self-Refine. In the final aggregation step of our method, we employ greedy decoding for closed-ended tasks to ensure deterministic outputs, and temperature sampling for open-ended tasks, following standard practice. For detailed prompts and parameter settings, please refer to Appendix \ref{exp-detail} and \ref{exp-prompt}.

\subsection{Main Results}

\begin{table*}[thb]
\centering
\small
\caption{Performance comparison on Llama 3 and GPT 4o mini across tasks, and methods. GSM8K, Math, GPQA, MMLU, SVAMP and MBPP scores are accuracy percentages. MT-bench is on a 1-10 scale; Alpaca scores are length-control win rate percentages against responses from GPT4-turbo; Best results for each model-task pair are in \textbf{bold}.}
\label{tab:performance}
\resizebox{\textwidth}{!}{%
\begin{tabular}{llccccc|ccc}
\toprule
\multirow{2}{*}{\textbf{Model}} & \multirow{2}{*}{\textbf{Method}} & \multicolumn{8}{c}{\textbf{Tasks}} \\
\cmidrule(lr){3-10}
& & GSM8K & MATH & GPQA & MMLU & SVAMP & \makecell{MT-bench} & \makecell{Alpaca} & MBPP \\
\midrule
\multirow{6}{*}{Llama 3} & Best-of-N (Oracle) & 91.74 & 47.26 & 61.23 & 77.08 & 92.67 & 8.04 & 36.14 & 60.60 \\
\cmidrule(lr){2-10}
& Greedy & 82.47 & 29.28 & 32.14 & 63.13 & 82.67 & 7.43 & 27.55 & \textbf{55.20} \\
& Self-Refine & 82.99 & 30.32 & 32.14 & 63.53 & 83.00 & 7.30 & 24.42 & 52.80 \\
& Self-Consistency & \textbf{86.35} & 31.68 & 33.26 & \textbf{65.62} & \textbf{88.33} & N/A & N/A & N/A \\
& Choose-from-N & 84.99 & 31.28 & 33.26 & 64.48 & 87.33 & 7.45 & 29.14 & 53.20 \\
& Ours & 86.05 & \textbf{32.46} & \textbf{35.04} & \textbf{65.62} & 88.00 & \textbf{7.53} & \textbf{29.34} & \textbf{55.20} \\
\midrule
\multirow{6}{*}{GPT 4o Mini} & Best-of-N (Oracle) & 96.29 & 86.72 & 56.47 & 86.25 & 95.33 & 9.30 & 61.67 & 78.60 \\
\cmidrule(lr){2-10}
& Greedy & 93.48 & 76.54 & 41.07 & 80.85 & 93.00 & 8.95 & 47.99 & 73.20 \\
& Self-Refine & 92.42 & 76.44 & 40.03 & 83.03 & 91.00 & 9.02 & 51.60 & 70.60 \\
& Self-Consistency & \textbf{94.77} & 77.26 & 40.40 & 85.00 & 93.00 & N/A & N/A & N/A \\
& Choose-from-N & 94.47 & 76.80 & 38.39 & 80.51 & \textbf{94.00} & 8.84 & 50.20 & 72.60 \\
& Ours & 94.69 & \textbf{78.25} & \textbf{42.17} & \textbf{85.11} & \textbf{94.00} & \textbf{9.13} & \textbf{55.85} & \textbf{74.20} \\
\bottomrule
\end{tabular}
}
\end{table*}

\begin{table*}[thb]
\centering
\vspace{-0.1in}
\small
\caption{Performance comparison on Gemma 2 and Qwen2.5 across tasks. GSM8K, GPQA and MBPP scores are accuracy percentages. MT-bench is on a 1-10 scale.}
\label{tab:empty_performance}
\begin{tabular}{llcccc}
\toprule
\multirow{2}{*}{\textbf{Model}} & \multirow{2}{*}{\textbf{Method}} & \multicolumn{4}{c}{\textbf{Tasks}} \\
\cmidrule(lr){3-6}
& & GSM8K & GPQA & \makecell{MT-bench} & MBPP \\
\midrule
\multirow{6}{*}{Gemma 2 9B} & Best-of-N (Oracle) & 93.03 & 50.45 & 8.74 &  61.80\\
\cmidrule(lr){2-6}
& Greedy & 87.11 & 31.74 & 7.99 & \textbf{57.60} \\
& Self-Refine & 87.04 & 31.70 & \textbf{8.35} & 57.40 \\
& Self-Consistency & 89.08 & 33.26 & N/A & N/A \\
& Choose-from-N & 88.25 & 31.92 & 8.30 & 57.20 \\
& Ours & \textbf{89.61} & \textbf{34.82} & \textbf{8.35} & 57.40 \\
\midrule
\multirow{6}{*}{Qwen 2.5 14B} & Best-of-N (Oracle) & 97.19 & 63.17 & 9.22 & 80.2 \\
\cmidrule(lr){2-6}
& Greedy & 94.62 & 39.51 & 8.63 &  72.00 \\
& Self-Refine & 94.62 & 40.18 & 8.65 &  71.80\\
& Self-Consistency & 95.68 & 40.40 & N/A &  N/A \\
& Choose-from-N & 95.68 & 40.62 & 8.66 &  73.00\\
& Ours & \textbf{95.75} & \textbf{41.74} & \textbf{8.99} &  \textbf{73.80}\\
\bottomrule
\end{tabular}
\end{table*}

Table \ref{tab:performance} presents comprehensive evaluation results across different models, tasks, and methods.
Our method outperforms choose-from-N and self-refine across different tasks and models, indicating that generative aggregation is more effective than model-based selection and feedback. Notably, on some tasks, choose-from-N and self-refine perform worse than greedy decoding (e.g., GPQA with GPT 4o Mini), which aligns with our hypothesis that the common language model training procedure does not adequately develop models' ability to make discriminative judgments.
The Best-of-N (Oracle) results reveal varying degrees of headroom for improvement across tasks, with our method's gains tending to be larger when Oracle performance significantly exceeds greedy decoding, suggesting that GSA benefits from the presence of diverse, high-quality candidates.

For mathematical reasoning and knowledge-based tasks, our approach also matches or outperforms Self-Consistency despite using fewer candidates (3 vs 4). For instance, with Llama 3, our method achieves comparable results on GSM8K and MMLU , while showing improvements on MATH and GPQA.
Beyond these structured tasks, our method also shows improvements on open-ended tasks where Self-Consistency is not applicable. With GPT4o Mini, we achieve notable gains over greedy decoding on Alpaca-eval and MT-bench. These improvements across different evaluation metrics and model scales demonstrate the broad applicability of our generative aggregation approach.

\begin{figure}[htb!]
\centering
\subfigure[Llama 3 8B]{\label{n_math}
\includegraphics[width=0.34\linewidth]{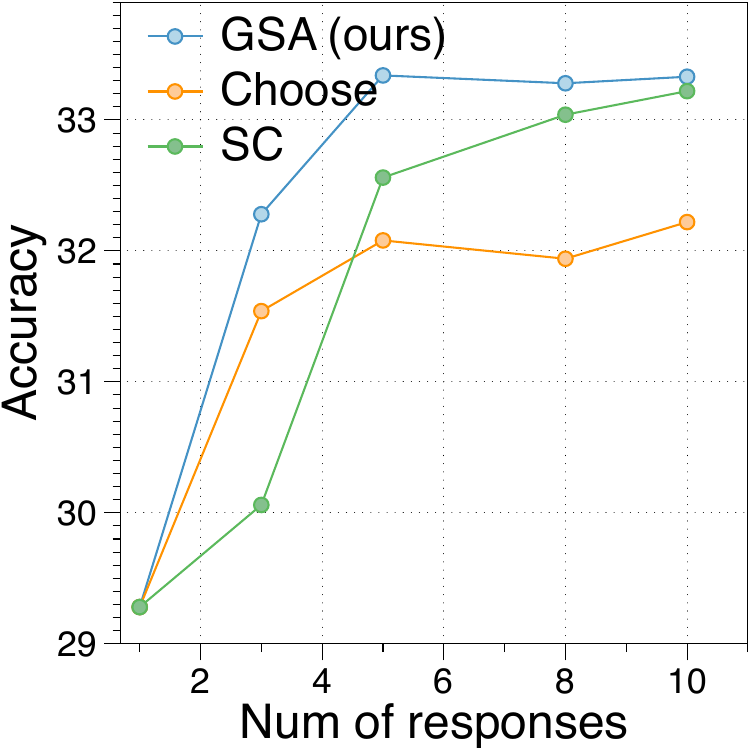}%
}
\subfigure[GPT 4o mini]{\label{n_math_gpt}
\includegraphics[width=0.34\linewidth]{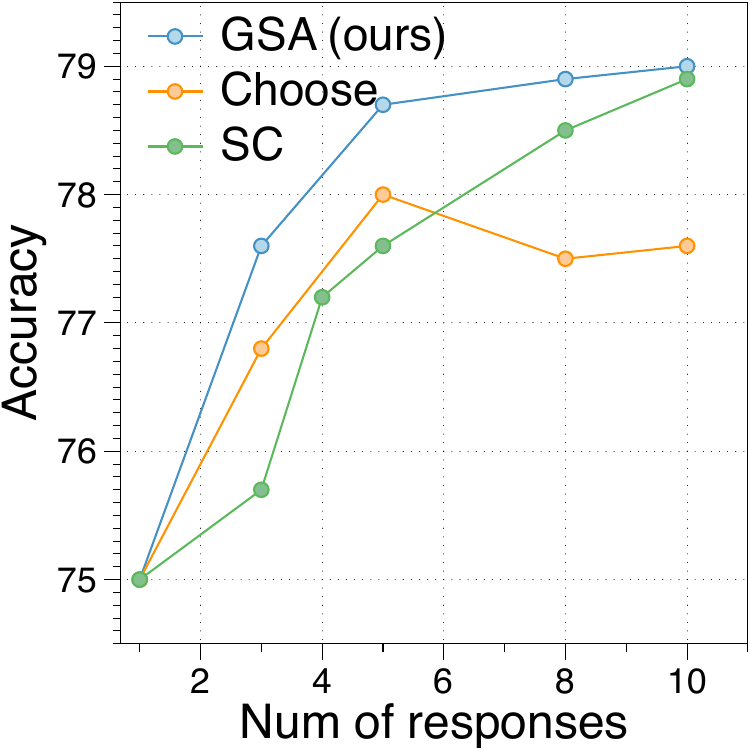}%
}
\vspace{-0.1in}
\caption{Comparison of baselines performance with different number of responses on MATH with Llama 3 8B and GPT 4o mini.}
\label{fig:N}
\vspace{-0.1in}
\end{figure}

\subsection{Ablation Studies}

\textbf{Number of responses.} We analyze how the number of responses ($N$) affects different aggregation methods' performance on MATH with Llama 3 8B and GPT 4o mini (tested on a 20\% subset due to API constraints). Note that here we apply same diverse responses for all baselines to compare the aggregation methods, with self-consistency requiring one less model call as it uses majority voting for aggregation. We apply this setting for the rest of ablation studies to better compare the aggregation methods. As shown in Figure \ref{fig:N}, all methods improve significantly when moving from greedy decoding ($N=1$) to multiple responses.
Our method consistently outperforms the choose-from-N baseline across different values of $N$, demonstrating the advantage of leveraging generative capabilities over attempting to select the best response. With GPT 4o mini, both our method and self-consistency continue to benefit from more responses, indicating their ability to effectively aggregate information from a larger set of examples, while the choose-from-N baseline shows a slight decrease at larger $N$, suggesting that selecting among many samples can be challenging. With Llama 3 8B, while self-consistency continues to benefit from more responses, the performance of our method peaks at $N=5$ and slightly decreases afterward, likely due to the model's difficulty in processing longer context when presented with too many responses.
\\

\noindent\textbf{Sampling temperatures.}
We investigate how sampling temperature affects the performance of different methods on GSM8K and GPQA with Llama 3 8B. All methods show a clear pattern where performance improves as temperature increases from 0.2 to 1.0, then degrades at higher temperatures. With low temperature, the responses lack diversity, resulting in similar performance across methods. The Oracle performance at this temperature also indicates limited diversity in the candidate pool. As temperature increases to 1.0, all methods benefit from increased response diversity, with our method achieve comparable performance to Self-consistency on GSM8K and better on GPQA, while consistently outperforming choose-from-N. Further increasing temperature leads to performance degradation due to the decreased quality of sampled responses.

\begin{figure}[htb!]
\centering
\subfigure[GSM8K]{\label{temp_gsm8k}
\includegraphics[width=0.34\linewidth]{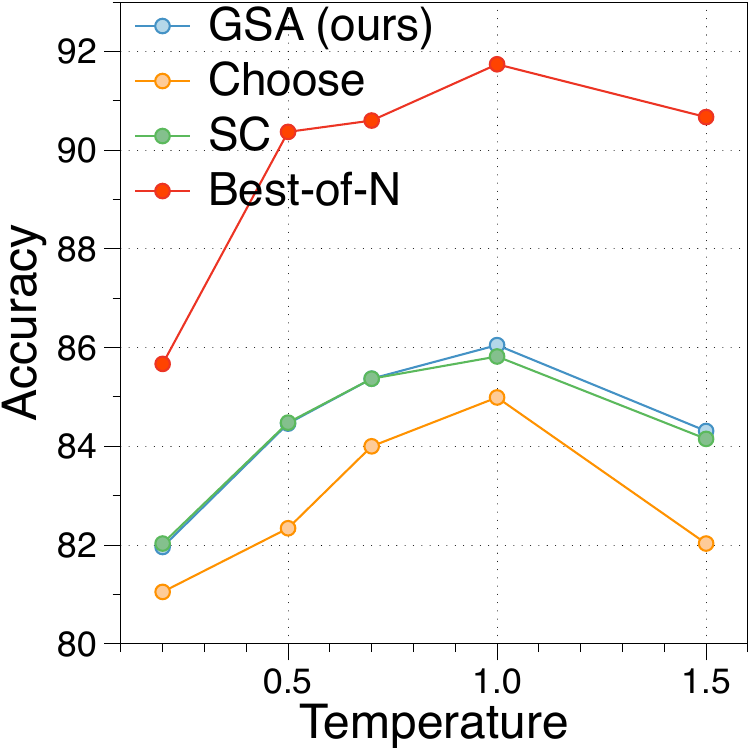}%
}
\subfigure[GPQA]{\label{temp_math}
\includegraphics[width=0.34\linewidth]{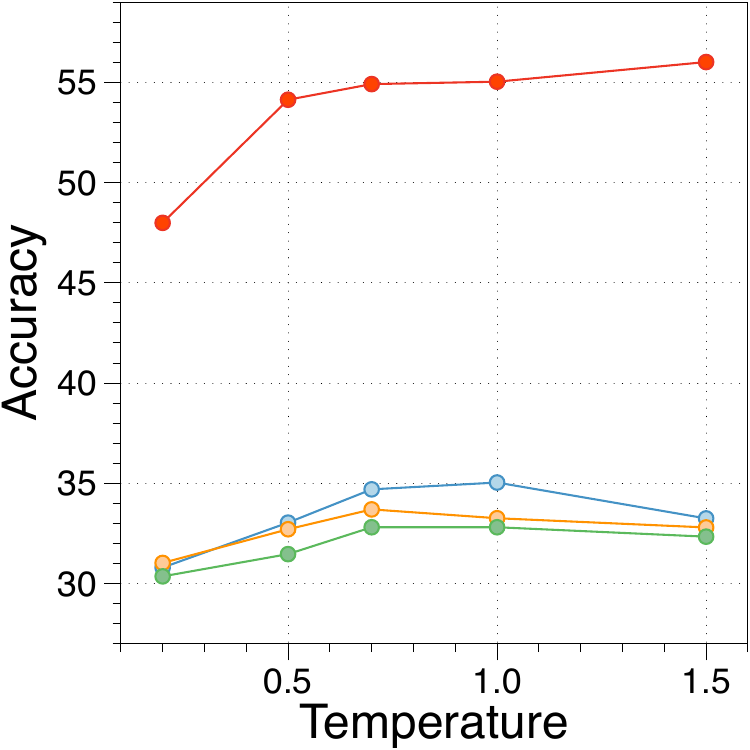}%
}
\vspace{-0.15in}
\caption{Comparison of baselines performance with different temperature and $N=3$ on GSM8K and GPQA with Llama 3 8B.}
\label{fig:temp}
\end{figure}

\noindent\textbf{Sampling strategies.}
Beyond temperature-based sampling, we explore two additional strategies for generating diverse responses: (1) \textbf{Prompt template variation} use different prompt templates to obtain diverse responses with distinct format. (2) \textbf{Multilingual generation} make use of the multilingual capabilities of model, prompting them to answer the question in different languages. Table \ref{tab:gemma_sampling} shows that all three sampling strategies achieve comparable performance on GSM8K using Gemma 2 9B with $N=3$. Our method maintains strong performance across all strategies, indicating our method's robustness to the choice of sampling method. 
The multilingual sampling leads to slightly better oracle and self-consistency performance, possibly because responses in different languages provide more diversity. However, it require multilingual capabilities of the LLM.

\begin{table}[htb!]
\centering
\small
\caption{Performance comparison of different sampling strategies on GSM8K for Gemma 2 9B with $N=3$. \textbf{Temp.} uses temperature sampling with $T=0.7$, \textbf{Prompt} uses different prompt templates, and \textbf{Multi.} generates responses in different languages.}
\label{tab:gemma_sampling}
\begin{tabular}{lccc}
\toprule
\textbf{Method} & \textbf{Temp.} & \textbf{Prompt} & \textbf{Multi.} \\
\midrule
Best-of-N (Oracle) & 92.49 & 92.95 & 92.80 \\
Self-Consistency & 88.48 & 88.86 & \textbf{89.84} \\
Choose-from-N & 88.25 & 88.17 & 89.16 \\
Ours & \textbf{89.61} & \textbf{89.23} & 89.31 \\
\bottomrule
\end{tabular}
\end{table}

\subsection{Discussions}

\begin{figure}
    \centering 
    \includegraphics[width=0.65\linewidth]{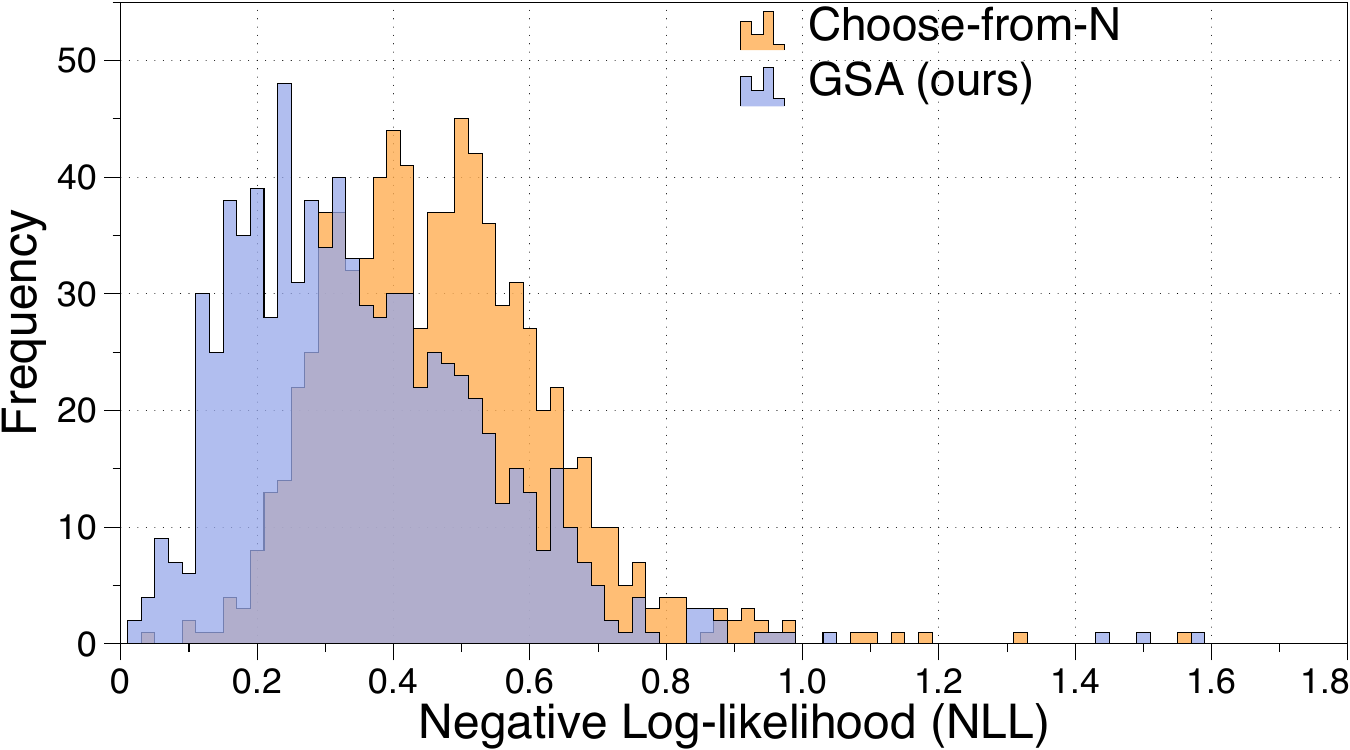}
    \caption{Distribution of normalized negative log-likelihood scores for responses generated on the Alpaca eval using Llama 3 8B. Lower NLL indicates higher model confidence in token generation.}
    \label{fig:nll} 
\vspace{-0.2in}
\end{figure}

\noindent\textbf{Likelihood distribution.} To further investigate our hypothesis that LLMs are better suited for generating new responses than discriminative task, we analyze the normalized negative log-likelihood (NLL) distributions of responses produced by our method and the choose-from-N baseline on the Alpaca eval benchmark using Llama 3 8B. Figure \ref{fig:nll} shows that our approach yields lower NLL values compared to choose-from-N, indicating higher model confidence when generating new responses than when making selections.

We note that while lower NLL values may correlate with improved response quality, they alone are insufficient for optimal response selection. Prior work \cite{selfconsistency} has shown that simply selecting responses with the lowest NLL values performs substantially worse than self-consistency. \\



\begin{figure}[htb!]
\centering
\subfigure[GPQA]{\label{overlap_gpqa}
\includegraphics[width=0.48\linewidth]{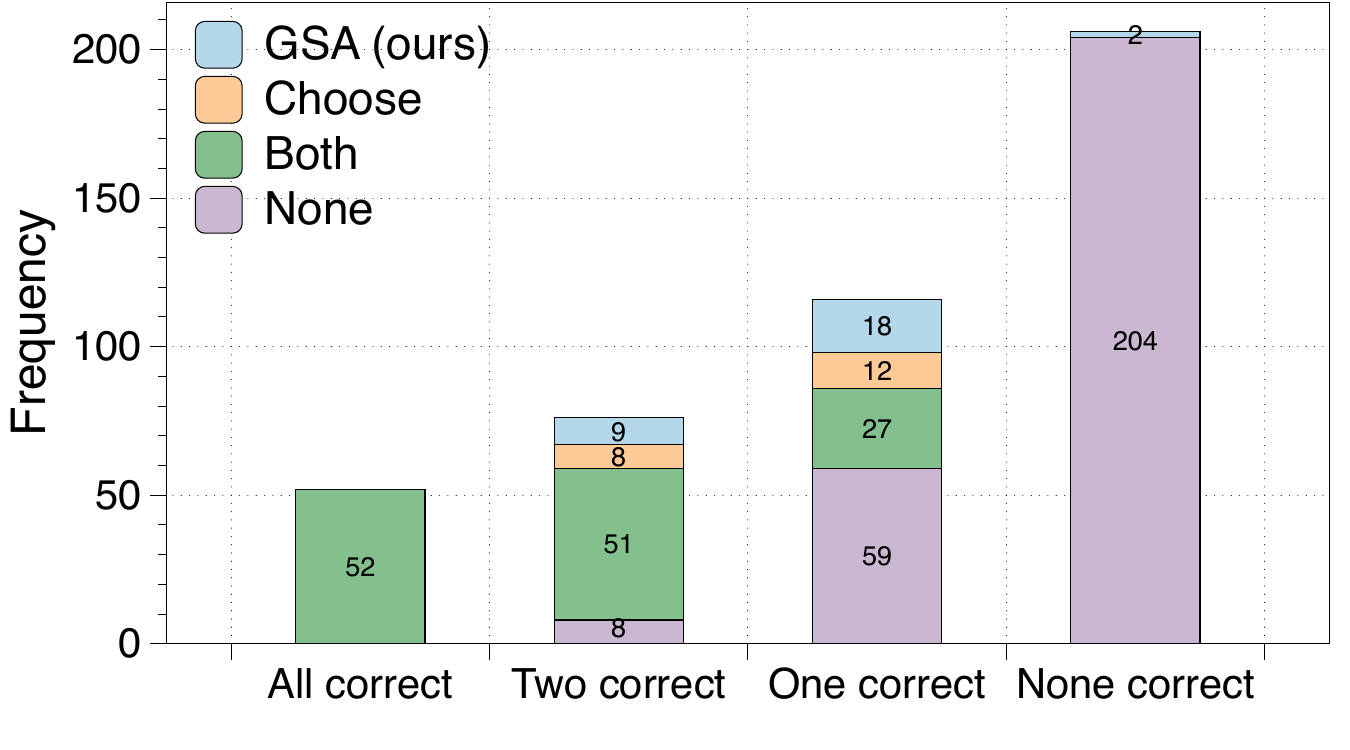}%
}
\subfigure[MBPP]{\label{overlap_mbpp}
\includegraphics[width=0.48\linewidth]{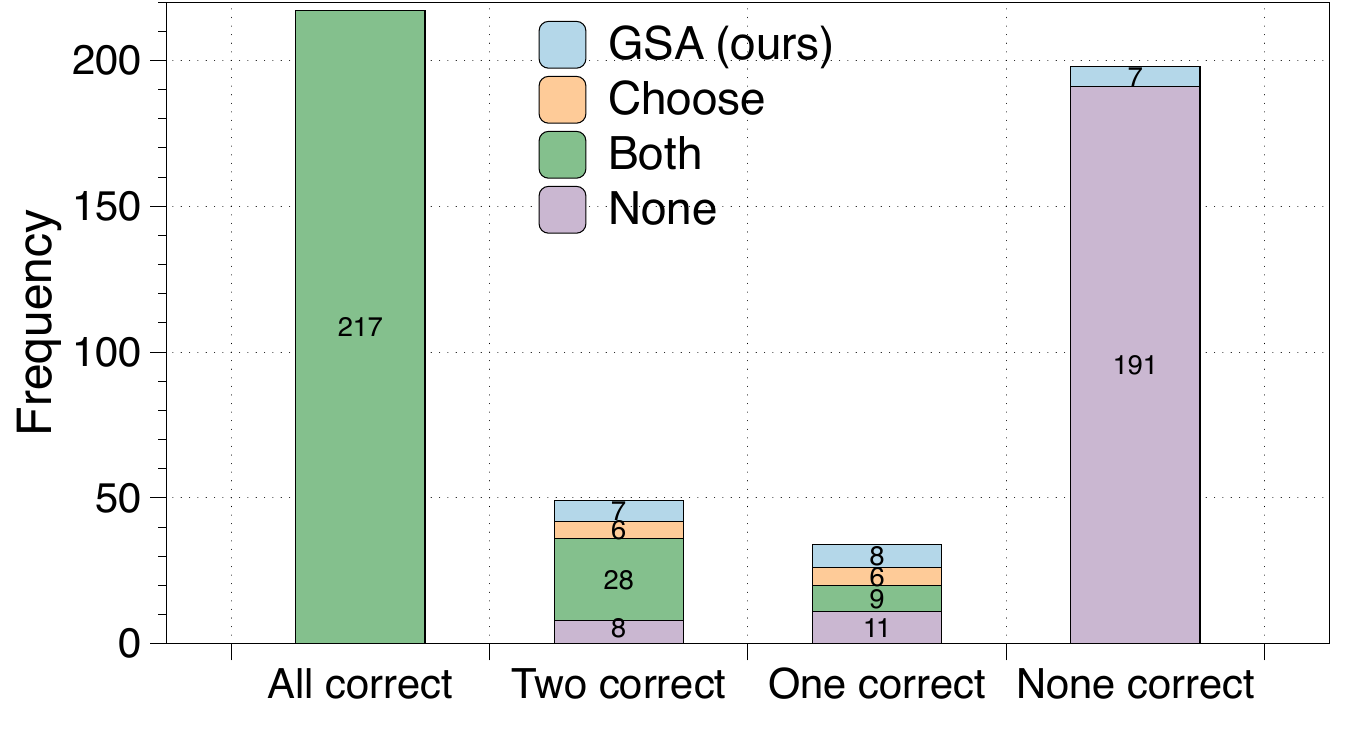}%
}
\caption{Detailed performance comparison between our method and choose-from-N on GPQA and MBPP tasks with Llama 3 8B. Test samples are grouped based on the number of correct responses among their three candidates. Within each group, we show the number of samples where: both methods, only our method, only choose-from-N, and neither method succeeds.}
\label{fig:overlap}
\vspace{-0.1in}
\end{figure}

\noindent\textbf{Fine-grained Performance Analysis} To gain deeper insights into how our method and choose-from-N utilize multiple candidate responses, we conduct a detailed comparative analysis on GPQA and MBPP with Llama 3 8B. We first categorize test samples based on the number of correct responses among the three candidates (3, 2, 1, or 0 correct). For each category, we then analyze four possible outcomes: both our method and choose-from-N succeed, only our method succeeds, only choose-from-N succeeds, or neither method succeeds. Figure \ref{fig:overlap} visualizes this breakdown.
When all three candidates are correct, both methods consistently produce correct answers. When two candidates are correct, both methods perform well, with our approach succeeding on slightly more samples than choose-from-N. The advantage of our approach becomes more significant when only one candidate is correct. Notably, our method can solves some cases where none of the original candidates were correct, demonstrating its ability to synthesize a correct solution even from incorrect examples. In contrast, choose-from-N is limited to selecting from existing candidates and hence cannot succeed when all candidates are incorrect.

\begin{figure*}[thb!]
\centering
\includegraphics[width=0.95\linewidth]{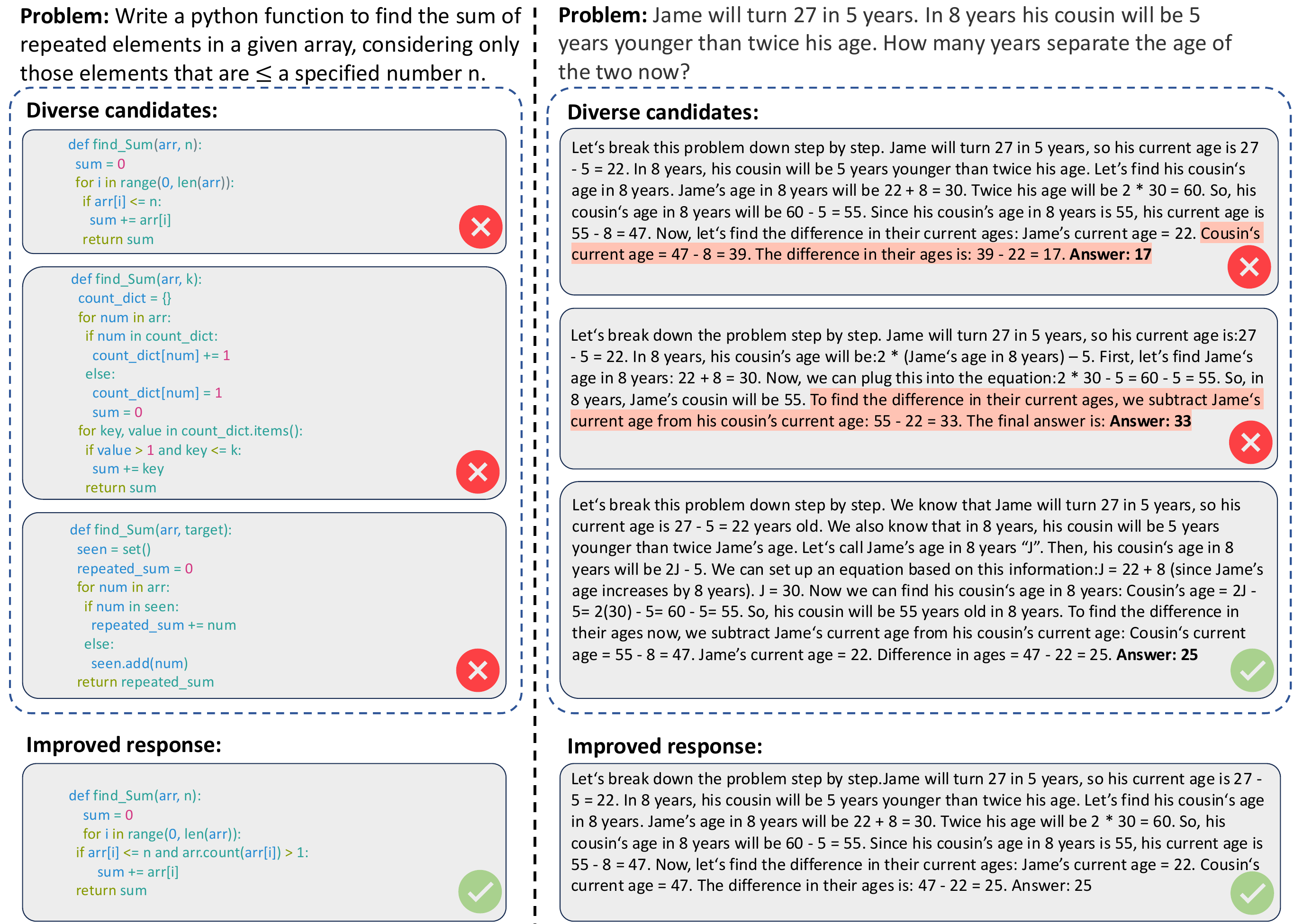}
\caption{Examples of successful GSA applications. Left: Combining different coding approaches to create a simpler and correct implementation. Right: Preserving clear reasoning structure while avoiding calculation errors.}
\label{fig:case}
\vspace{-0.1in}
\end{figure*}

\subsection{Case Study}

We present two examples on coding and mathematical reasoning tasks to provide insights into how our method operates. As shown in Figure \ref{fig:case}, we demonstrate how GSA effectively aggregates information from diverse candidate solutions to generate improved responses.
In the coding task, the diverse candidates showcase different strategies, each with distinct limitations.
Our method aggregates these approaches into an improved solution that combines simplicity of the first solution and the counting in the second.
In the math task, while all candidates correctly determine that Jame's cousin's age in 8 years, two responses make errors in calculating the current age difference.
The third response reaches the correct answer but employs a less intuitive algebraic approach using variables.
Our method's response preserves the more straightforward calculation path seen in the first two responses, and correctly derive the final answer.

\section{Conclusion}

In this paper, we introduced Generative Self-Aggregation (GSA), a novel prompting method that improves LLMs' performance without relying on discriminative judgments. Our approach demonstrates that LLMs can effectively aggregate information from multiple solutions through generative processes, utilizing diverse reasoning paths to produce enhanced responses. GSA requires neither additional training nor external feedback, making it readily applicable across different model architectures and domains. Our extensive empirical evaluation across diverse tasks demonstrates that GSA outperforms existing self-correction and choose-from-N methods that rely on LLMs' discriminative capabilities. As future work, GSA could be used to generate high-quality supervised data for model fine-tuning, and specialized training focused on enhancing LLMs' aggregation capabilities may further improve performance.

\section*{Impact Statement}
This paper presents work whose goal is to advance the field of Machine Learning. There are many potential societal consequences of our work, none of which we feel must be specifically highlighted here.

\section*{Acknowledgement}
All experiments were conducted at Georgia Tech and Microsoft.

\bibliography{reference}
\bibliographystyle{ims}

\newpage
\appendix
\onecolumn
\section{Implementation Details and Parameters}
\label{exp-detail}

We conduct all experiments using inference-only settings, utilizing A6000 40GB GPUs and vLLM for efficient inference with open-source models, while accessing GPT-4o-mini through the OpenAI API service. For mathematical reasoning tasks, we set the maximum new token length to 2048, extending it to 4096 for GPQA and MMLU. When evaluating open-ended tasks (MT-bench, Alpaca eval, MBPP), we adhere to the default settings specified by each benchmark's evaluation convention. For all open-source models, we maintain a consistent top-p value of 0.95 during inference.

The temperature settings for candidate generation are tuned based on empirical performance of self-consistency and our method across different models and tasks. For LLaMA-3, we employ temperatures of 1.0 for GSM8K and GPQA, 0.7 for MATH, SVAMP, MMLU, and MT-bench, 0.5 for MTbench, and 0.8 for Alpaca eval. GPT-4o-mini uses a consistent temperature of 0.7 across all tasks except Alpaca eval, where we set it to 1.0. Gemma-2 maintains a uniform temperature of 0.7. Qwen-2.5 uses task-specific settings: 0.7 for GSM8K and MT-bench, 1.0 for GPQA, and 0.5 for MBPP. For the aggregation step, we employ greedy decoding (temperature = 0) for closed-ended tasks while maintaining task-specific temperature settings for open-ended benchmarks.

\section{Full Sets Of Prompts}
\label{exp-prompt}
We list the full details of the prompts used for candidates generation, choose-from-N and our method on each task. We apply zero-shot setting for all baselines and hence do not require constructing any few-shot examples.

\subsection{GSM8K and SVAMP}
\begin{lstlisting}[language={}, caption={Candidates generation prompt for GSM8K and SVAMP},  breakindent=0pt]
Question: {question}
Please put the final answer at the end of your response in the format "Answer: <number>". Let's solve this step by step:
\end{lstlisting}

\begin{lstlisting}[language=, caption={Aggregation prompt for GSM8K and SVAMP}, breakindent=0pt]
Question: {question}
Here are some potential responses:
{responses_text}

Given these solutions, please consider their consistency, and please provide a correct solution to the question with clear reasoning and step-by-step calculations.
Please put the final answer at the end of your response in the format "Answer: <number>".
\end{lstlisting}

\begin{lstlisting}[language=, caption={Choose-from-N prompt for GSM8K and SVAMP},  breakindent=0pt]
Question: {question}
Here are some potential responses:
{responses_text}

Given these solutions, please consider their consistency and choose a correct one. Give me clear explanation of your choice and put the index of the correct answer at the end of the response. Please put the index in the format "Index: <index>". The index should be in the range of 1 to {num_responses}.
\end{lstlisting}

For prompting variation in ablation study, we use the following prompt:
\begin{lstlisting}[language=, caption={Prompt variation for GSM8K}, breakindent=0pt]
# Prompt 1
Question: {question}
Please put the final answer at the end of your response in the format "Answer: <number>". Let's solve this step by step:

# Prompt 2
Question: {question}
Imagine you are explaining this problem to a student learning math for the first time. Be clear and concise, and end your explanation with "Answer: <number>".

# Prompt 3
Question: {question}
Solve the problem step by step, checking for potential errors along the way. Provide the final answer at the end: "Answer: <number>".

# Multilingual
[{language}] Question: {question}
Please put the final answer at the end of your response in the format "Answer: <number>". Let's solve this step by step using {language}:
\end{lstlisting}

\subsection{MATH}

\begin{lstlisting}[language={}, caption={Candidates generation prompt for MATH},  breakindent=0pt]
Question: {question}
Please put the final answer at the end of your response in the form of \\boxed{...}. Let's solve this step by step:
\end{lstlisting}

\begin{lstlisting}[language=, caption={Aggregation prompt for MATH}, breakindent=0pt]
Question: {question}
Here are some potential responses:
{responses_text}

Given these solutions, please consider their consistency, and please provide a correct solution to the question with clear reasoning and step-by-step calculations.
Please put the final answer at the end of your response in the form of \\boxed{...}.
\end{lstlisting}

\begin{lstlisting}[language=, caption={Choose-from-N prompt for MATH},  breakindent=0pt]
Question: {question}
Here are some potential responses:
{responses_text}

Given these solutions, please consider their consistency and choose a correct one. Give me clear explanation of your choice and put the index of the correct answer at the end of the response. Please put the index in the format "Index: <index>". The index should be in the range of 1 to {num_responses}.
\end{lstlisting}

\subsection{GPQA}

\begin{lstlisting}[language={}, caption={Candidates generation prompt for GPQA},  breakindent=0pt]
Question: {question}
Choices: {choices}

Please select an answer for the question from the above choices. Put the final answer as a **single letter** at the end of the response in the format "The correct answer is (insert answer here)". Let's think step by step:
\end{lstlisting}

\begin{lstlisting}[language=, caption={Aggregation prompt for MMLU}, breakindent=0pt]
{question and choices}

Here are some potential responses:
{responses_text}

Given these solutions, please analyze their consistency and correctness, and then provide a correct solution with clear reasoning. 
Put the final answer as a single letter at the end of your response in the format "The correct answer is (insert answer here)".
\end{lstlisting}

\begin{lstlisting}[language=, caption={Choose-from-N prompt for GPQA},  breakindent=0pt]
{question and choices}

Here are some potential responses:
{responses_text}

Given these solutions, please consider their consistency and choose a correct one. Give me clear explanation of your choice and put the index (1-{num_responses}) of the correct answer at the end of the response.
Put the index of the correct answer as a single number in the format "The correct index is (insert index here)".
\end{lstlisting}

\subsection{MMLU}

\begin{lstlisting}[language={}, caption={Candidates generation prompt for MMLU},  breakindent=0pt]
Question: {question}
Choices: {choices}

Please select an answer for the question from the above choices. Put the final answer as a **single letter** at the end of the response in the format "The correct answer is (insert answer here)". Let's think step by step:
\end{lstlisting}

\begin{lstlisting}[language=, caption={Aggregation prompt for MMLU}, breakindent=0pt]
{question and choices}

Here are some potential responses:
{responses_text}

Please review the given solutions, and then provide a correct answer with clear reasoning. 
Put the final answer as a single letter at the end of your response in the format "The correct answer is (insert answer here)".
\end{lstlisting}

\begin{lstlisting}[language=, caption={Choose-from-N prompt for MMLU},  breakindent=0pt]
{question and choices}

Here are some potential responses:
{responses_text}

Please review the given solutions, and then give me the index (1-{num_responses}) of the correct answer at the end of the response.
Put the index of the correct answer as a single number in the format "The correct index is (insert index here)".
\end{lstlisting}

\subsection{MT-bench}

\begin{lstlisting}[language=, caption={Aggregation prompt for MT-bench}, breakindent=0pt]
{query}
Below are some responses to this instruction:
{responses_text}

Please review the above responses and generate a better response to the instruction: <{query}>.
\end{lstlisting}

\begin{lstlisting}[language=, caption={Choose-from-N prompt for MT-bench},  breakindent=0pt]
{query}
Below are some responses to this instruction:
{responses_text}

Please review the above responses and choose a best response by providing the index (1-{n_responses}) of the best response. Please put the index at the end of your response in the format "Index: <number>"."""
\end{lstlisting}

\subsection{Alpaca Eval}

\begin{lstlisting}[language=, caption={Aggregation prompt for Alpaca Eval}, breakindent=0pt]
###Instruction:
1) **Review** the following problem and the reference solutions provided.
2) **Provide** your own answer to the problem.
3) **Provide** a brief explanation of your reasoning.

###Reference Solutions:
{references_text}
###Input:
Here is the problem:
{question}
\end{lstlisting}

\begin{lstlisting}[language=, caption={Choose-from-N prompt for Alpaca Eval},  breakindent=0pt]
### Instruction:
1) **Review** the following problem and the reference solutions provided.
2) **Provide only** the index number of the best solution of the correct solution in the format "Index: <number>". 
3) **Provide** a brief explanation of your reasoning.

### Reference Solutions:
{solutions_text}
### Input:
Here is the problem:
{question}
\end{lstlisting}

\subsection{MBPP}

\begin{lstlisting}[language=, caption={Aggregation prompt for MT-bench}, breakindent=0pt]
Here is the problem:
{prompt}
### Reference Solutions:
{references_text}
### Instructions:
1. Review the above solutions.
2. **Generate** a Python function that solves the Problem.
3. **Provide** a brief explanation of your reasoning.
4. **Ensure** your code is enclosed within a ```python``` code block.
\end{lstlisting}

\begin{lstlisting}[language=, caption={Choose-from-N prompt for MT-bench},  breakindent=0pt]
### Instruction:
1) **Review** the following problem and the reference solutions provided.
2) **Provide only** the index number of the best solution of the correct solution in the format "Index: <number>". 
3) **Provide** a brief explanation of your reasoning.

### Reference Solutions:
{solutions_text}
### Input:
Here is the problem:
{prompt}

\end{lstlisting}


\end{document}